\documentclass[10pt,journal,compsoc]{IEEEtran}
\ifCLASSOPTIONcompsoc
  \usepackage[nocompress]{cite}
\else
  \usepackage{cite}
\fi
\ifCLASSINFOpdf
\else
\fi

\usepackage{bm}
\usepackage{amsmath}
\usepackage{subfigure}
\usepackage{array,graphicx}
\usepackage{subfloat}
\usepackage{graphicx}
\usepackage{cite}
\usepackage{algorithm}
\usepackage{multirow}
\usepackage[square, comma, sort&compress, numbers]{natbib}
\usepackage{mathrsfs}
\usepackage{caption2}
\usepackage{color}
\usepackage{booktabs}
\usepackage{times}
\usepackage{epsfig}
\usepackage{graphicx}
\usepackage{amsmath}
\usepackage{amssymb}
\usepackage{bbm}
\usepackage{multirow}
\usepackage{eqparbox,array}
\usepackage{makecell}
\usepackage{tabularx}
\usepackage{multirow}
\usepackage{placeins}
\usepackage{float}
\usepackage[T1]{fontenc}

\begin{document}
%
\title{ConsistentID\includegraphics[width=1.0cm]{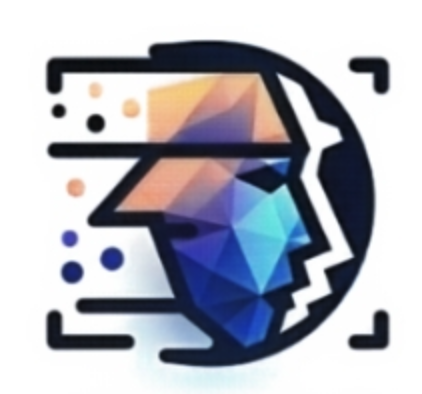}: Portrait Generation with Multimodal Fine-Grained Identity Preserving}
%
%
%
%

\author{Jiehui~Huang, \quad 
        Xiao~Dong, \quad
        Wenhui~Song,\quad
        Zheng~Chong,\quad
        Zhenchao~Tang,\quad
        Jun~Zhou,\quad
        Yuhao~Cheng,\quad
        Long~Chen,\quad
        Hanhui~Li,\quad
        Yiqiang~Yan,\quad
        Shengcai~Liao,\quad
        and\quad~Xiaodan~Liang
 \IEEEcompsocitemizethanks{\IEEEcompsocthanksitem 
J.H. Huang, W.H. Song, Z. Chong, Z.C. Tang, J. Zhou, H.H. Li, X.D. Liang are with the School of Artificial Intelligence,  Shenzhen Campus, Sun Yat-Sen University, Shenzhen, P.R. China, 518107. 
X. Dong with the School of Intelligent Systems Engineering, Zhuhai Campus, Sun Yat-Sen University, Zhuhai,  P.R. China, 519082. 
Y.H. Cheng, L. Chen, Y.Q. Yan are with the Lenovo Research Group, Shenzhen, P.R. China, 518038.  
S.C. Liao is with College of Information Technology, United Arab Emirates University, Al Ain, UAE.

E-mail: {(jhhuang117@gmail.com, huangjh336@mail2.sysu.edu.cn); (dx.icandoit@gmail.com, dongx55@mails2.sysu.edu.cn); songwh6@mail2.sysu.edu.cn; chongzh@mail2.sysu.edu.cn; tangzhch7@mail2.sysu.edu.cn; zhouj235@mail2.sysu.edu.cn; chengyh5@Lenovo.com; chenlong12@lenovo.com; lihanhui@mail3.sysu.edu.cn; yanyq@lenovo.com; scliao@ieee.org; xdliang328@gmail.com}

}
\thanks{Xiaodan Liang is the Corresponding Author.}
}
\IEEEtitleabstractindextext{%
\begin{abstract} 
Diffusion-based technologies have made significant strides, particularly in personalized and customized facial generation.  
However, existing methods struggle to achieve high-fidelity and detailed identity (ID) consistency.  
This is mainly due to two challenges: insufficient fine-grained control over specific facial areas and the absence of a comprehensive strategy for ID preservation that accounts for both intricate facial details and the overall facial structure. 
To address these limitations, we introduce ConsistentID, an innovative method crafted for diverse identity-preserving portrait generation under fine-grained multimodal facial prompts, utilizing only a single reference image.  
ConsistentID comprises two core components: a multimodal facial prompt generator and an ID-preservation network.  
The facial prompt generator combines localized facial features, facial feature descriptions, and overall facial descriptions to enhance the precision of facial detail reconstruction.  
The ID-preservation network, optimized with a facial attention localization strategy, ensures consistent identity preservation across facial regions.  
Together, these components leverage fine-grained multimodal identity information to improve identity preservation accuracy significantly. 
To drive ConsistentID's training, we propose a fine-grained portrait dataset, FGID, with over 500,000 facial images, offering greater diversity and comprehensiveness than existing public facial datasets.
Experimental results substantiate that our ConsistentID achieves exceptional precision and diversity in personalized facial generation, surpassing existing methods in the MyStyle dataset. 
In addition, although ConsistentID introduces more multimodal ID information, it still maintains rapid inference speed during the generation process.  
Our codes and pre-trained checkpoints are available at
https://github.com/JackAILab/ConsistentID.
\end{abstract}

}

\maketitle

\IEEEdisplaynontitleabstractindextext

%
\IEEEpeerreviewmaketitle

\IEEEraisesectionheading{\section{Introduction}\label{sec:intro}}

\IEEEPARstart{I}{-}mage-generation technology \citep{ju2023humansd, liu2023hyperhuman, ruiz2023hyperdreambooth, huang2023collaborative, stypulkowski2024diffused} has undergone significant evolution, driven by the emergence and advancement of diffusion-based~\citep{song2020denoising,ho2020denoising} text-to-image large models like GLIDE~\citep{nichol2021glide}, DALL-E 2~\citep{ramesh2022hierarchical}, Imagen~\citep{saharia2022photorealistic}, Stable Diffusion (SD)~\citep{rombach2022high}, eDiff-I~\citep{balaji2022ediffi} and RAPHAEL~\citep{xue2024raphael}. 
This progress has given rise to a multitude of application approaches across diverse scenarios.  
Among these, personalized portrait generation has garnered significant attention in both academia and industry due to its wide-ranging applications, including e-commerce advertising, personalized gift customization, and virtual try-ons.  

The primary challenge in customized facial generation lies in maintaining facial image consistency across different attributes based on one or multiple reference images, leading to two key issues: ensuring accurate identity (ID) consistency and achieving high-fidelity, diverse facial details.  
Current text-to-image models~\citep{zhang2023adding, ye2023ip, ruiz2023hyperdreambooth, wang2018toward, valevski2023face0}, despite incorporating structural and content guidance, face limitations in accurately controlling personalized and customized generation, particularly concerning the fidelity of generated images to reference images.

To improve the precision and diversity of personalized portrait generation with reference images, numerous customized methodologies have emerged, meeting users' demands for high-quality customized images.  
These personalized approaches are categorized based on whether fine-tuning occurs during inference, resulting in two distinct types: test-time fine-tuning and direct inference.  
\textbf{Test-time fine-tuning}:  This category includes methods such as Textual Inversion \citep{gal2022image}, HyperDreambooth\citep{ruiz2023hyperdreambooth}, and CustomDiffusion\citep{kumari2023multi}. 
Users can achieve personalized generation by providing a set of target ID images for post-training. 
Despite achieving commendable high-fidelity results, the quality of the generated output depends on the quality of manually collected data.  
Additionally, the manual collection of customized data for fine-tuning introduces a labor-intensive and time-consuming aspect, limiting its practicality.
\textbf{Direct inference}: Another category of models, including IP-Adapter~\citep{ye2023ip}, Fastcomposer\citep{xiao2023fastcomposer}, Photomaker~\citep{li2023photomaker}, and InstantID~\citep{wang2024instantid}, adopts a single-stage inference approach.  
These models enhance global ID consistency by either utilizing the image as a conditional input or manipulating image-trigger words. 
 However, most methods frequently overlook fine-grained information, such as landmarks and facial features.  
Although InstantID~\citep{wang2024instantid} enhances ID consistency to some extent by incorporating landmarks, the use of visual prompt landmarks limits the diversity and flexibility of key facial regions, resulting in rigidly generated facial features. 
Building on these observations, it becomes evident that \textbf{two pivotal challenges} persist in personalized portrait generation, requiring meticulous consideration:  1) neglect of fine-grained facial information and 2) identity inconsistency between facial areas and the whole face, as illustrated in Figure~\ref{fig:facial_compare}. 

\begin{figure*}[h]
    \centering
    \includegraphics[width=1.0\textwidth]{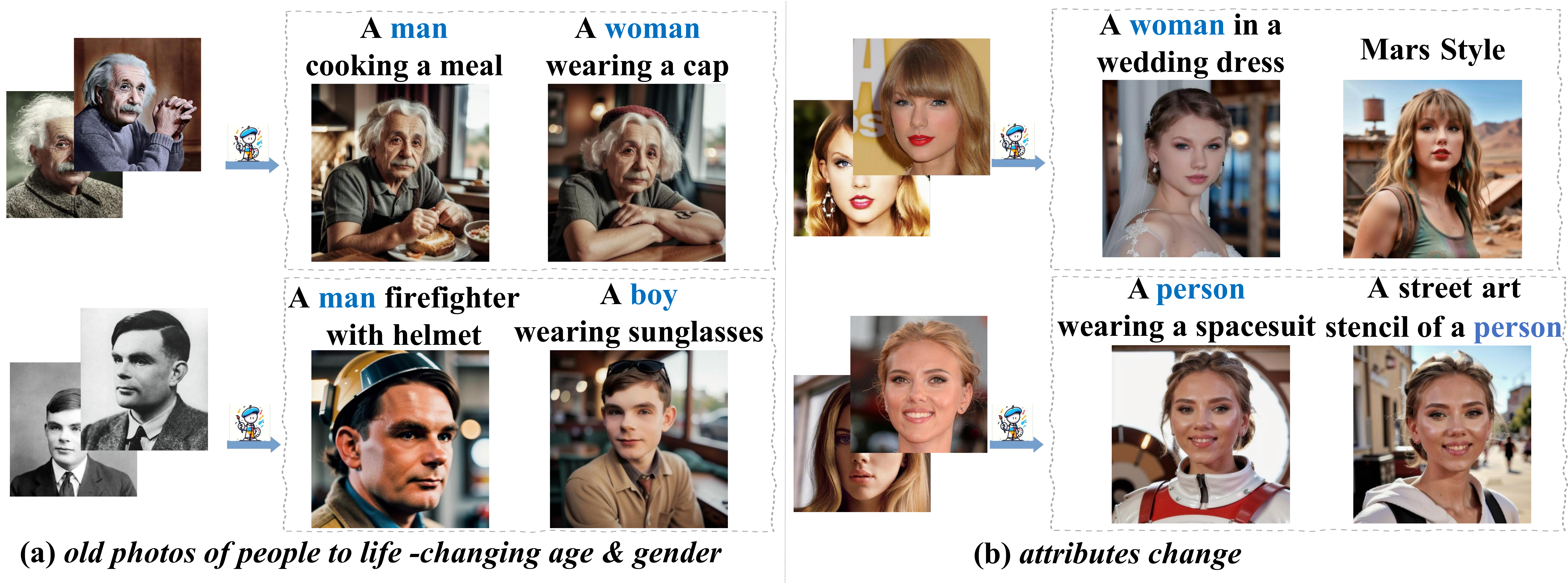}
    \caption{
    Given some images of input IDs, our ConsistentID can generate diverse personalized ID images based on text prompts using only a single image.}
    \label{fig:teaser}
\end{figure*}


To tackle identity consistency and detail preservation challenges in personalized image generation, we propose \textbf{ConsistentID}, a novel method that excels in maintaining identity fidelity while capturing diverse facial details. 
With just a single facial image as input, our approach leverages multimodal fine-grained ID features to deliver high-quality results. 
As illustrated in Figure~\ref{fig:teaser}, ConsistentID enables personalized transformations such as revitalizing old photos, altering gender, or changing outfits.  
By seamlessly integrating facial inputs with text prompts, our model generates diverse yet consistent identity representations, offering unmatched adaptability in personalized generation tasks. 

\begin{figure*}[tb]
     \includegraphics[width=1\textwidth]{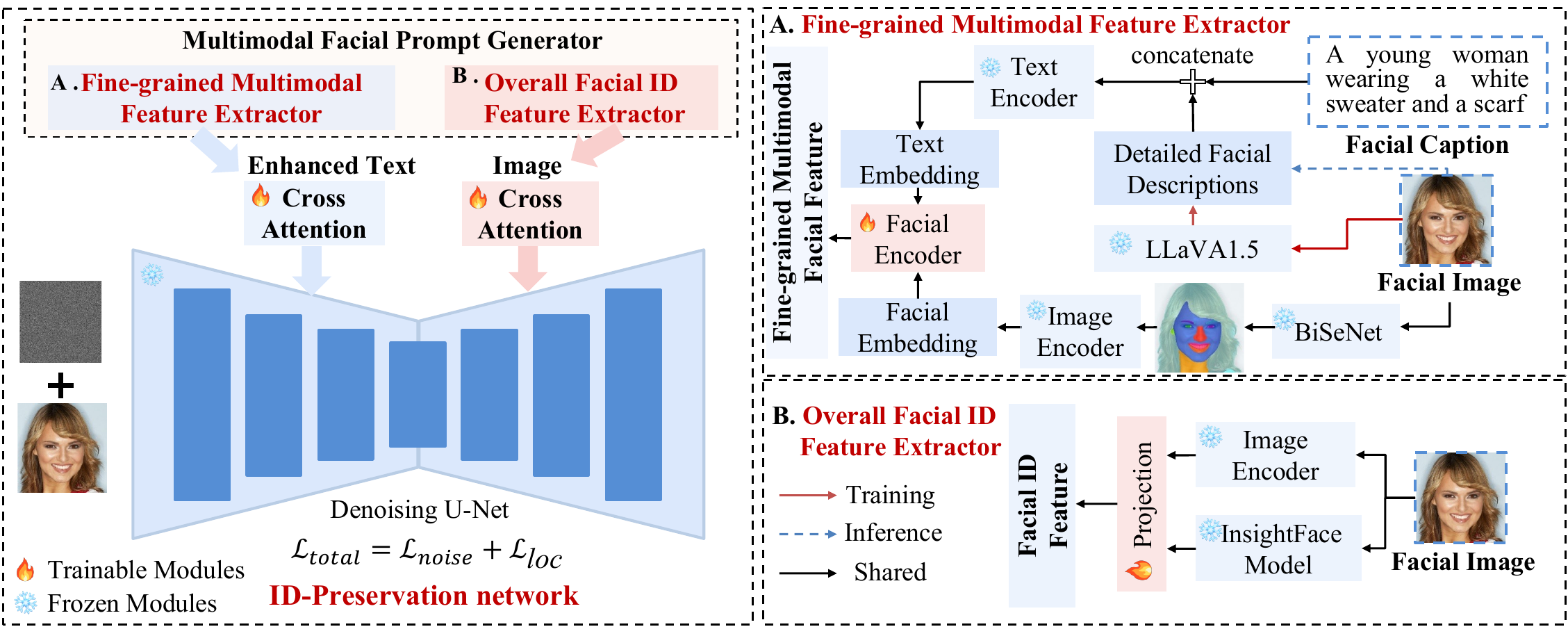}
    \caption{The overall framework of our proposed ConsistentID. The framework comprises two key modules: a multimodal facial ID generator and a purposefully crafted ID-preservation network. The multimodal facial prompt generator consists of two essential components: a fine-grained multimodal feature extractor, which focuses on capturing detailed facial information, and a facial ID feature extractor dedicated to learning facial ID features. On the other hand, the ID-preservation network utilizes both facial textual and visual prompts, preventing the blending of ID information from different facial regions through the facial attention localization strategy. This approach ensures the preservation of ID consistency in the facial regions. }
        \label{fig:framework}
\end{figure*}

Figure~\ref{fig:framework} provides an overview of our ConsistentID. 
ConsistentID comprises two key modules: 1) a multimodal facial prompt generator and 2) an ID-preservation network. 
The former component includes a fine-grained multimodal feature extractor and a facial ID feature extractor, enabling the generation of more detailed facial ID features using multi-conditions, incorporating facial images, facial regions, and corresponding textual descriptions extracted from the multimodal large language model LLaVA1.5~\citep{liu2023improved}.  
The facial ID features generated by the first module are then passed to the latter module, which enhances ID consistency across facial regions through a facial attention localization strategy. 
Additionally, we recognize the limitations of existing portrait datasets~\citep{zheng2022general,cao2018vggface2,liu2015deep,nitzan2022mystyle,wang2020mead}, particularly in capturing diverse and fine-grained facial details that preserve identity, crucial to the effectiveness of ConsistentID.  
To address this, we introduce the inaugural Fine-Grained ID Preservation (FGID) dataset, along with a fine-grained identity consistency metric, providing a unique and comprehensive evaluation approach to enhance our training and provide a comprehensive evaluation of its performance in capturing facial details. 

In summary, our contributions are as follows. 
\begin{itemize}
\item [$ \bullet $]  We introduce ConsistentID to improve fine-grained customized facial generation by incorporating detailed descriptions of facial regions and local facial features. 
The experimental results demonstrate the superiority of ConsistentID in terms of ID consistency and high fidelity, even with only one reference image. 
Additionally, although ConsistentID introduced more detailed multimodal fine-grained ID information during training,  it achieves this with a single fixed prompt~\footnote{This person has one nose, two eyes, two ears, and a mouth.} during inference.
It does not depend on facial descriptions generated by LLaVA1.5, ensuring a streamlined and efficient approach, as shown in Table~\ref{tab:my_label}. 
\item [$ \bullet $]  We devise an ID-preservation network optimized by facial attention localization strategy, enabling more accurate ID preservation and more vivid facial generation. 
This mechanism ensures the preservation of ID consistency within each facial region by preventing the blending of ID information from different facial regions.
\item [$ \bullet $] 

We introduce the inaugural fine-grained facial generation dataset, FGID, addressing limitations in existing datasets for capturing diverse identity-preserving facial details.  
This dataset includes facial features and descriptions of both facial regions and the entire face, complemented by a novel fine-grained identity consistency metric, establishing a comprehensive evaluation framework for fine-grained facial generation performance.
\end{itemize}

\section{Related Work}\label{sec:related_work}
\label{sec:related_work}
\noindent\textbf{Text-to-image Diffusion Models.} Diffusion models have made notable advancements, garnering significant attention from both industry and academia, primarily due to their exceptional semantic precision and high fidelity.  
The success of these models can be attributed to the utilization of high-quality image-text datasets, continual refinement of foundation modality encoders, and the iterative enhancement of controlled modules. 
In the domain of text-to-image generation, the encoding of text prompts involves utilizing a pretrained language encoder, such as CLIP~\citep{radford2021learning}, to transform it into a latent representation, subsequently inserted into the diffusion model through the cross-attention mechanism. 
Pioneering models in this domain include GLIDE~\citep{nichol2021glide}, SD~\citep{rombach2022high}, DiT~\citep{peebles2023scalable}, among others, and further developments and innovations are continuing to emerge~\citep{huang2023composer,xue2024raphael}.  
A notable advancement in this lineage is SDXL~\citep{podell2023sdxl}, which stands out as the most powerful text-to-image generation model. 
It incorporates a larger Unet~\citep{ronneberger2015u} model and employs two text encoders for enhanced semantic control and refinement.  
As a follow-up, we use the SD~\citep{rombach2022high} model as our base model to achieve personalized portrait generation. 

\noindent\textbf{Personalization in Diffusion Models.} 
Due to the potent generative capability of the text-to-image diffusion model, many personalized generation models are constructed based on it.  
The mainstream personalized image synthesis methods are categorized into two groups based on whether fine-tuning occurs during test time.  
One group relies on optimization during test-time, with typical methods including Dreambooth~\citep{ruiz2023dreambooth}, Textual Inversion~\citep{gal2022image}, IP-Adapter~\citep{wang2024instantid}, ControlNet~\citep{zhang2023adding}, Custom Diffusion~\citep{kumari2023multi}, and LoRA~\citep{hu2021lora}. Dreambooth and Textual Inversion fine-tune a special token S* to learn the concept during the fine-tuning stage.  
In contrast, IP-Adapter, ControlNet, and LoRA insert image semantics using an additional learned module, such as cross-attention, to imbue a pre-trained model with visual reasoning understanding ability.  

Despite their advancements, these methods necessitate resource-intensive backpropagation during each iteration, making the learning process time-consuming and limiting their practicality.    
Recently, researchers have focused more on methods bypassing additional fine-tuning or inversion processes, mainly including IP-Adapter~\citep{ye2023ip}, FastComposer~\citep{xiao2023fastcomposer}, PhotoMaker~\citep{li2023photomaker}, and InstantID~\citep{wang2024instantid}.  
This type of method performs personalized generation using only an image with a single forward process, which is more advantageous in calculation efficiency compared to the former type.  
However, we observe that fine-grained facial features are not fully considered in the training process, easily leading to ID inconsistency or lower image quality, as shown in Figure~\ref{fig:result_compare}.  
To address these limitations, we introduce ConsistentID, aiming to mitigate the ID-preserving issue and enhance fine-grained control capabilities while reducing data dependency.  
Our approach incorporates a specially designed facial encoder using detailed descriptions of facial features and local image conditions as inputs.   
Additionally, we contribute to a new landmark in the facial generation field by proposing: 1) the introduction of the first fine-grained facial generation datasets and 2) the presentation of a new metric that redefines the performance evaluation of facial generation.  

\section{Method}
In this section, we will introduce our multimodal facial prompt generator in Subsection~\ref{subsec_generator_1} and propose a meticulously designed ID-preservation network in Subsection~\ref{subsec_network}.  
Next, we detail the training process of ConsistentID and its inference procedure in Subsection~\ref{subsec_training_infer}. 
Finally, we provide a comprehensive description of the FGID dataset in Subsection~\ref{sub:sec_dataset}.  
\label{sec_model}

\subsection{Multimodal Facial Prompt Generator}
\label{subsec_generator_1}
\noindent
\noindent \textbf{Fine-grained Multimodal Feature Extractor.} In this module, we independently learn fine-grained facial visual and textual embeddings and feed them into the designed lightweight facial encoder to generate fine-grained multi-modal facial features. 
Three key components are used in the module, including text embedding, facial embedding and facial encoder.

\noindent  1) \textbf{Text Embedding.}
Motivated by recent works~\citep{liu2023fine, liu2023hyperhuman, nasir2019text2facegan, wan2018fine} in personalized facial generation, our goal is to introduce more detailed and accurate facial descriptions.   
To achieve this, the entire facial image is processed by the Multimodal Large Language Model (MLLM) LLaVA1.5~\citep{liu2023improved} using the prompt: `Describe this person's facial features, including the face, ears, eyes, nose, and mouth'.  
The mentioned process will generate feature-level descriptions of the facial areas. 
Next, the terms `face', `ears', `eyes', `nose', and `mouth' in these descriptions are replaced with the delimiter `$<$facial$>$', and the modified text is concatenated with captions describing the entire facial image. 
Finally, the concatenated descriptions are input into a pre-trained text encoder to learn the fine-grained text embedding of the facial features. 
With the abundant descriptions from facial regions, the text embeddings are enriched with more precise identity information, effectively resolving identity inconsistency.

\noindent 2) \textbf{Facial Embedding.} 
In contrast to existing methods~\citep{meng2021sdedit, karras2019style}, \citep{meng2021sdedit, karras2019style,gal2022image, gu2024photoswap, liu2024towards, rosberg2023facedancer,xiao2023fastcomposer, li2023photomaker, wang2024instantid} that rely on brief textual descriptions or coarse-grained visual prompts, our goal is to integrate more fine-grained multimodal control information at the facial region level, aiming to achieve greater accuracy in capturing detailed facial features. 
To enrich the ID-preservation information, we delve into more fine-grained facial features, including eye gaze, earlobe characteristics, nose shape, and others.  
Following the previous method~\citep{huang2020interpretable, umirzakova2022detailed, yu2023towards, li2023mask, yu2023freedom, kim2022diffface}, we employ the pre-trained face model BiSeNet~\citep{yu2018bisenet} to extract segmentation masks of facial areas, encompassing eyes, nose, ears, mouth, and other regions, from the entire face.  
Subsequently, the facial regions obtained from these masks are fed into the pre-trained image encoder to learn fine-grained facial embeddings. 
The inclusion of facial regions' features results in fine-grained facial embeddings containing more abundant ID-preservation information compared to features learned from the entire face.  

\noindent 3) \textbf{Facial Encoder.} Previous studies~\citep{radford2021learning, xiao2023fastcomposer, gal2022image, ruiz2023dreambooth, yu2018bisenet} have demonstrated that relying solely on visual or textual prompts cannot comprehensively maintain ID consistency both in appearance and semantic details.   
While IP-Adapter~\citep{ye2023ip} makes the initial attempt to simultaneously inject multimodal information through two distinct decoupled cross-attention mechanisms, it overlooks ID information from crucial facial regions, rendering it susceptible to ID inconsistency in facial details. 

\begin{figure*}[t!]
  \centering  
  \includegraphics[width=1\textwidth]{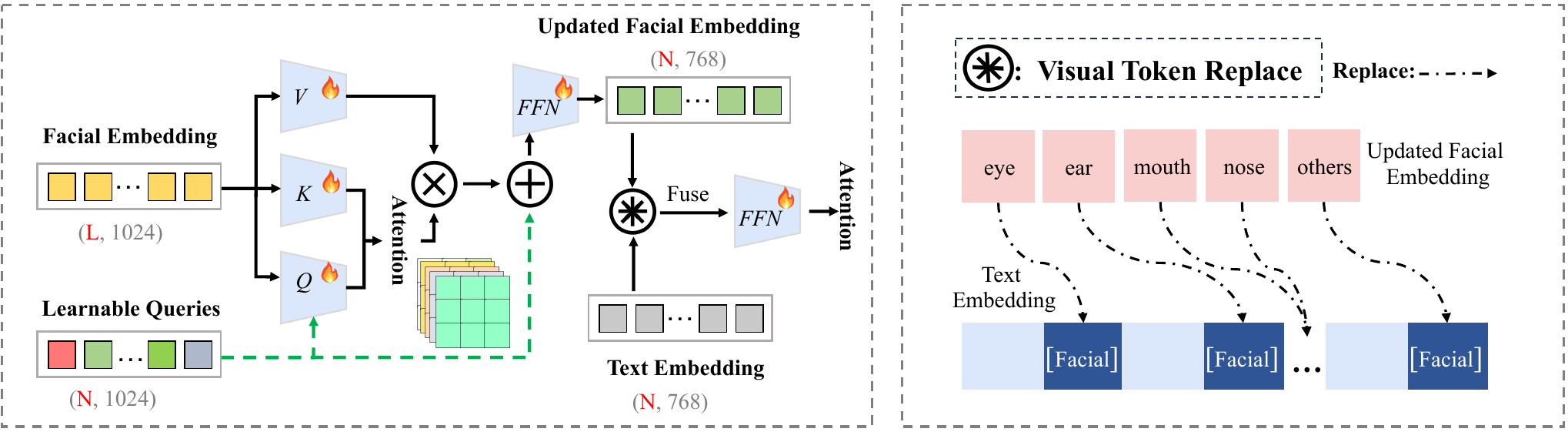} 
  \caption{The framework of our facial encoder for generating fine-grained multimodal facial features.
  }
  \label{fig:facial_encoder}  
\end{figure*}

To cultivate the potential of image and text prompts, inspired by the token fusion approach of multimodal large language models, we design a facial encoder to seamlessly integrate visual prompts with text prompts along the dimension of the text sequence, as depicted in Figure~\ref{fig:facial_encoder}.  
Specifically, given a facial embedding and a caption embedding, the facial encoder initially employs a self-attention mechanism to align the entire facial features with facial areas' features, resulting in aligned features denoted as ${\widehat{f}^{i} \in \mathbb{R}^{N \times D}}$, where $N=5$ represents the number of facial feature areas, including eyes, mouth, ears, nose, and other facial regions, and $D$ represents the dimension of text embeddings.  
In cases where face images lack a complete set of $N$ facial features, the missing features are padded using an all-zero matrix.  
Subsequently, we replace the text features at the position of the delimiter `$<$facial$>$' with $\widehat{f}^{i}$ using the visual token replacement operation, as illustrated in Figure~\ref{fig:facial_encoder} (right). 
Finally, the text features, now enriched with visual identity information, are fed into two multi-layer perceptions (MLP) to learn the text conditional embeddings.

\noindent \textbf{Facial ID Feature Extractor.} 
Except for the input condition of fine-grained facial features, we also inject the character's overall ID information into our ConsistentID as a visual prompt.  
This process relies on the pre-trained CLIP image encoder and the pre-trained face model from the specialized version of the IP-Adapter~\citep{ye2023ip} model, IPA-FaceID-Plus~\citep{ye2023ip}. 
Specifically, the complete facial images are simultaneously fed into both encoders for visual feature extraction. 
Following these two encoders, a lightweight projection module, with parameters initialized by IPA-FaceID-Plus, is used to generate the face embedding of the whole image.  

\subsection{ID-Preservation Network}
\label{subsec_network}
The effectiveness of visual prompts in pre-trained text-to-image diffusion models~\citep{xiao2023fastcomposer,li2023photomaker, zhang2023adding, chefer2023attend, qiu2024controlling, tumanyan2023plug} significantly enhances textual prompts, especially for content that is challenging to describe textually.  
However, visual prompts alone often provide only coarse-grained control due to the semantic fuzziness of visual tokens. 
To solve this, we integrate fine-grained multimodal ID prompts and overall ID prompts into the UNet model through the cross-attention module to achieve precise ID preservation.  

Specifically, we first introduced an ID consistency network, which maintains the consistency of local ID features by guiding the attention of facial features to align with the corresponding facial regions~\citep{xiao2023fastcomposer}.

This optimization strategy is derived from the observation that traditional cross-attention maps tend to simultaneously focus on the entire image rather than specific facial regions, making it challenging to maintain ID features during each facial region generation.  
To address this issue, we introduce facial segmentation masks during training to obtain attention scores learned from the enhanced text cross-attention module for facial regions.

Let \(P \in [0, 1]^{h \times w \times n}\) represent the cross-attention map that connects latent pixels to multimodal conditional embeddings at each layer, where \(P[i, j, k]\) signifies the attention map from the \(k\)-th conditional token to the (\(i\), \(j\)) latent pixel.  
Ideally, the attention maps of facial region tokens should focus exclusively on facial feature areas, preventing the blending of identities between facial features and averting propagation to the entire facial image.  
To achieve this goal,  we propose localizing the focus of the cross-attention map by using segmentation masks that correspond to the reference facial regional features.

Let \(M = \{m_1, m_2, m_3, ..., m_N\}\) represent the segmentation masks of the reference facial regions, \(I = \{i_1, i_2, i_3, ..., i_N\}\) as the index list indicating which facial feature corresponds to visual and textual tokens in the multimodal prompt, and \(P_i = P[:, :, i] \in [0, 1]^{h \times w}\) denote the cross-attention map of the \(i\)-th facial region's token, where \(I\) is generated using the special token `$<$facial$>$'.  

Given the cross-attention map \(P_{i_j}\), it should closely correspond to the facial region identified by the \(j\)-th multimodal token, segmented by \(m_{j}\).  
To achieve this, we introduce \(m_{j}\) and apply it to \(P_{i_j}\) to obtain its corresponding activation region, aligning with the segmentation mask \(m_{j}\) of the \(j\)-th facial feature token. 
For achieving this correspondence, a balanced $\mathcal{L}_1$ loss is employed to minimize the distance between the cross-attention map and the segmentation mask:
\begin{equation}
    \label{eq:loc}
    \mathcal{L}_{\text {facial}}=\frac{1}{N} \sum_{j=1}^{N}\left(\operatorname{mean}\left(P_{i_j}\left[1-m_j\right]\right)-\operatorname{mean}\left(P_{i_j}\left[m_j\right]\right)\right),
\end{equation}
where N denotes the number of the segmentation masks. 
This loss formulation aims to ensure that each facial feature token's attention map aligns closely with its corresponding segmentation mask, promoting precise and localized attention during the generation process.

\subsection{Training and Inference Details}
\label{subsec_training_infer}
The training data for ConsistentID consists of facial image-text pairs.   
In ConsistentID, to enhance text controllability, we prioritize the caption as the primary prompt and concatenate it with more detailed descriptions of facial regions extracted from LLaVA1.5, forming the ultimate textual input.  
During training, only the parameters of the facial encoder and projection module within the facial ID feature extractor are optimized, while the pre-trained diffusion model remains frozen.  
Regarding training loss functions, they align with those used in the original stable diffusion models and are expressed as:
\begin{equation}
    \mathcal{L}_{noise}=\mathbb{E}_{z_t, t, C_f, C_l, \epsilon \sim \mathcal{N}(0,1)}\left[\left\|\epsilon-\epsilon_\theta\left(z_t, t, C_f, C_i\right)\right\|_2^2\right],
\end{equation}
\noindent where \(C_l\) denotes the facial ID feature, and \(C_f\) is the fine-grained multimodal facial feature.  

The total loss function is $\mathcal{L}_{total} =  \mathcal{L}_{noise}+ \lambda \mathcal{L}_{\text {facial}}$.

During the inference process, we employ a straightforward delayed primacy condition as similar to Fastcomposer.  
This allows the use of a separate text representation initially, followed by enhanced text representation after a specific step, effectively balancing identity preservation and editability.  

\subsection{FGID Dataset}
\label{sub:sec_dataset}
Our ConsistentID necessitates detailed facial features and corresponding textual prompts to address issues like deformation, distortion, and blurring prevalent in current facial generation methods.  
However, existing datasets~\citep{karras2019style, david_beniaguev_2022_SFHQ, zheng2022general, chen2024subject} predominantly focus on local facial areas and lack fine-grained ID annotations for specific features such as the nose, mouth, eyes, and ears.

To address this limitation, we introduce the FGID dataset, which provides comprehensive fine-grained ID information and detailed facial descriptions essential for training the ConsistentID model. 
In the following, we will describe the dataset curation process and key characteristics of this dataset. 

\subsubsection{Dataset Curation}
\begin{figure}[!t]
    \centering
    \includegraphics[width=1.0\linewidth]{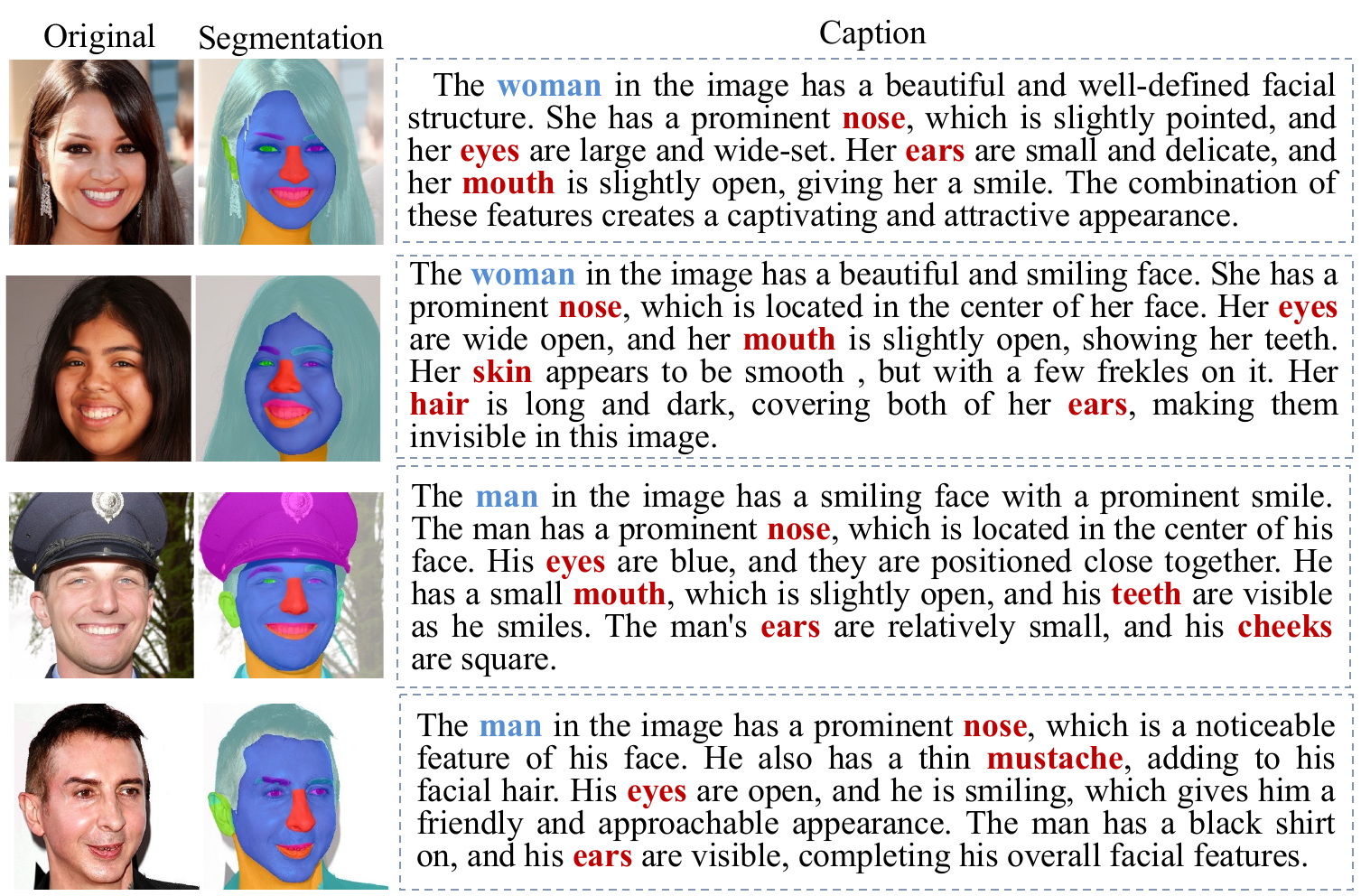}
    \caption{Several training data demos from our FGID dataset.}
    \label{fig:Caption_Segmentation}
\end{figure}

\noindent\textbf{Data source:} Our facial images in the FGID dataset are from three public datasets, including FFHQ~\citep{karras2019style}, CelebA~\citep{liu2015deep}, and SFHQ~\citep{david_beniaguev_2022_SFHQ}. 
We separately select 70,000, 30,000, and 424,258 images from these datasets.  
Finally, a total of 524,258 images are selected, where 107,048 images have recognizable IDs.  
Figure \ref{fig:Caption_Segmentation} shows some examples of these images and their corresponding captions. 

\noindent \textbf{Dataset Pipeline:} As for the selected images,  each image is first processed through BiSeNet~\citep{yu2018bisenet}  to generate a fine-grained binary mask (\(I_{\text{mask}}\)) that identifies predefined facial components. 
Using \(I_{\text{mask}}\), the corresponding facial regions are segmented from the original image, denoted as \(I_{\text{face}}\).  
Meanwhile, the InsightFace~\citep{deng2019arcface} model is used to extract comprehensive facial identity features, ensuring both global and regional identity information is captured effectively. 

For textual data, the LLaVA1.5 model generates detailed descriptions using the embedded prompt:  
`\textbf{Please describe the people in the image, including their gender, age, clothing, facial expressions, and any other distinguishing features.}'.
This ensures fine-grained textual annotations are paired with the visual data.

\subsubsection{Dataset Characteristics}

\begin{figure}
    \centering
    \includegraphics[width=1\linewidth]{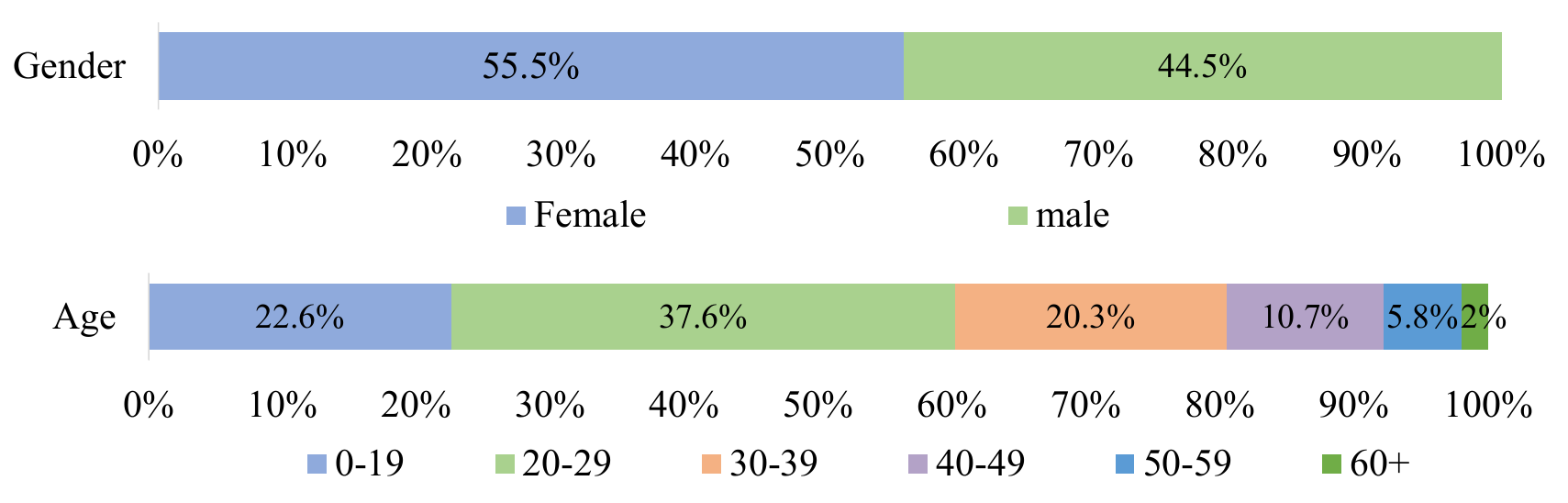}
    \caption{The statistical characteristics of age and gender distribution in the FGID training dataset.}
    \label{fig:fig_app_age}
\end{figure}

The FGID dataset is designed to provide comprehensive textual and visual information for training models like ConsistentID. Below, we summarize its key features and characteristics:

\noindent \textbf{Dataset Composition:}
The FGID dataset includes 15 distinct identity types, as detailed in Table~\ref{tab:identity_names}. To ensure diversity and inclusiveness, the distributions of gender and age across all identities are presented in Figure~\ref{fig:fig_app_age}, demonstrating a relatively balanced representation of these properties.

\noindent \textbf{Dataset Scenarios:}
The dataset spans 45 scenarios, categorized into four application areas: Clothing \& Accessory, Action, Background, and Style. Table~\ref{tab:prompta} lists the prompts used to define these scenarios, ensuring a wide range of contextual diversity. 

\noindent \textbf{Rich Data Representation:}
The FGID dataset is tailored to capture both whole-face and specific facial feature information, providing fine-grained textual and visual details essential for model training. This design enhances its utility for generating accurate and diverse facial representations.

\noindent \textbf{Future Directions:}
To further augment diversity and information content, we plan to expand the dataset by incorporating additional resources, such as higher-level datasets like LAION-Face~\citep{zheng2022general} and self-collected multi-ID data.

\begin{table}[tb]
    \centering
    \vspace{-2mm}
    \begin{tabular}{lll}
\hline  
    \multicolumn{3}{c}{Evaluation IDs} \\
\hline
   {1.} Andrew Ng & {6.} Scarlett Johansson & {11.} Joe Biden \\
   {2.} Barack Obama & {7.} Taylor Swift & {12.} Kamala Harris \\
   {3.} Dwayne Johnson  & {8} Albert Einstein & {13.} Kaming He \\
   {4.} Fei-Fei Li & {9.} Elon Mask & {14.} 
 Lecun Yann \\
   {5.} Michelle Obama & {10.} Geoffrey Hinton & {15.} Sam Altman \\
\hline
    \end{tabular}
    \caption{ID names used for evaluation.}
    \label{tab:identity_names}
\end{table}

\begin{table}[tb]
\centering
\resizebox{\columnwidth}{!}{
    \begin{tabularx}{\columnwidth}{l|X} 
    \hline
    Category & Prompt \\ 
    \hline

    \multirow{10}{*}{\makecell{Clothing\& \\Accessory}} 
    & a $<$class word$>$ wearing a red hat \\
    & a $<$class word$>$ wearing a santa hat \\
    & a $<$class word$>$ wearing a rainbow scarf \\
    & a $<$class word$>$ wearing a black top hat and a monocle \\
    & a $<$class word$>$ in a chef outfit \\
    & a $<$class word$>$ in a firefighter outfit \\
    & a $<$class word$>$ in a police outfit \\
    & a $<$class word$>$ wearing pink glasses \\
    & a $<$class word$>$ wearing a yellow shirt \\
    & a $<$class word$>$ in a purple wizard outfit \\
    \hline

    \multirow{10}{*}{Background} 
    & a $<$class word$>$ in the jungle \\
    & a $<$class word$>$ in the snow \\
    & a $<$class word$>$ on the beach \\
    & a $<$class word$>$ on a cobblestone street \\
    & a $<$class word$>$ on top of pink fabric \\
    & a $<$class word$>$ on top of a wooden floor \\
    & a $<$class word$>$ with a city in the background \\
    & a $<$class word$>$ with a mountain in the background \\
    & a $<$class word$>$ with a blue house in the background \\
    & a $<$class word$>$ on top of a purple rug in a forest \\
    \hline 

    \multirow{15}{*}{Action} 
    & a $<$class word$>$ holding a glass of wine \\
    & a $<$class word$>$ riding a horse \\
    & a $<$class word$>$ holding a piece of cake \\
    & a $<$class word$>$ giving a lecture \\
    & a $<$class word$>$ reading a book \\
    & a $<$class word$>$ gardening in the backyard \\
    & a $<$class word$>$ cooking a meal \\
    & a $<$class word$>$ working out at the gym \\
    & a $<$class word$>$ walking the dog \\
    & a $<$class word$>$ baking cookies \\
    & a $<$class word$>$ wearing a doctoral cap \\
    & a $<$class word$>$ wearing a spacesuit \\
    & a $<$class word$>$ wearing sunglasses and necklace \\
    & a $<$class word$>$ coding in front of a computer \\
    & a $<$class word$>$ in a helmet and vest riding a motorcycle \\
    \hline 

    \multirow{10}{*}{Style} 
    & a painting of a $<$class word$>$ in the style of Banksy \\
    & a painting of a $<$class word$>$ in the style of Vincent Van Gogh \\
    & a colorful graffiti painting of a $<$class word$>$ \\
    & a watercolor painting of a $<$class word$>$ \\
    & a Greek marble sculpture of a $<$class word$>$ \\
    & a street art mural of a $<$class word$>$ \\
    & a black and white photograph of a $<$class word$>$ \\
    & a pointillism painting of a $<$class word$>$ \\
    & a Japanese woodblock print of a $<$class word$>$ \\
    & a street art stencil of a $<$class word$>$ \\
    \hline
    \end{tabularx}
}
\caption{Evaluation text prompts are categorized by Clothing\&Accessories, Background, Action, and Style.  
During inference, the term `class' will be substituted with `man', `woman`, `girl`, etc. 
For each identity, we use these prompts to generate 45 images for evaluation.}
\label{tab:prompta}
\end{table}
\section{Experiments}
\label{sec:exp}

\begin{figure}[tb]
  \centering  
  \includegraphics[width=0.5\textwidth]{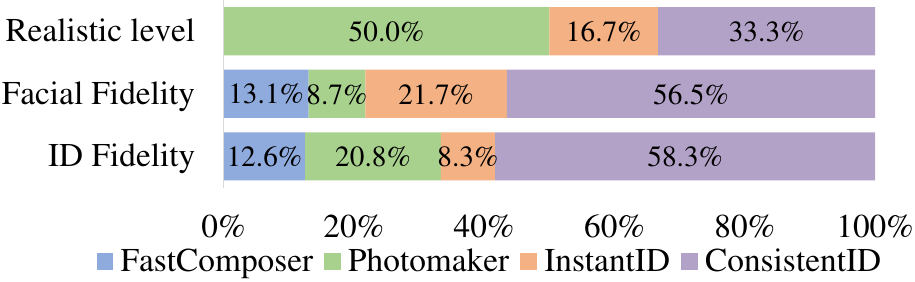}
  \caption{User preferences across image fidelity, fine-grained ID fidelity, overall ID fidelity for different methods.}
  \label{fig:user_pref}  
\end{figure}

\subsection{Implementation Details}
\label{experiment}
\noindent \textbf{Experimental Implements.} In ConsistentID, we employ the Stable Diffusion V1.5 (SD1.5) as the foundational text-to-image training model.  
For the fine-grained multimodal feature extractor, we initialize the parameters of all text encoders and image encoders with CLIP-ViT-H ~\citep{schuhmann2022laion}. 
Additionally, we use the image projection layers from CLIP-ViT-H to initialize the learnable projection module within the overall facial ID feature extractor. 
The entire framework is optimized using Adam~\citep{kingma2014adam} on 8 NVIDIA 3090 GPUs, with a batch size of 16.  
We set the learning rate for all trainable modules to  \(1 \times 10^{-4}\). 
During training, we probabilistically remove 50\% of the background information from the characters with a 50\% probability to mitigate interference.    
The hyperparameter coefficient ($\lambda$) of facial features loss $\mathcal{L}_{\text {facial}}$ is set to 0.01. 
Additionally, to enhance generation performance through classifier guidance, there is a 10\% chance of replacing the original updated text embedding with a zero text embedding. 
During inference, we use delayed topic conditioning ~\citep{karras2019style, xiao2023fastcomposer} to resolve text and ID condition conflicts. 
We generate portrait images using a 50-step DDIM sampler~\citep{song2020denoising} with a classifier guidance scale set to 5. 

\begin{figure*}[tb]
  \centering  
  \includegraphics[width=1.0\textwidth]{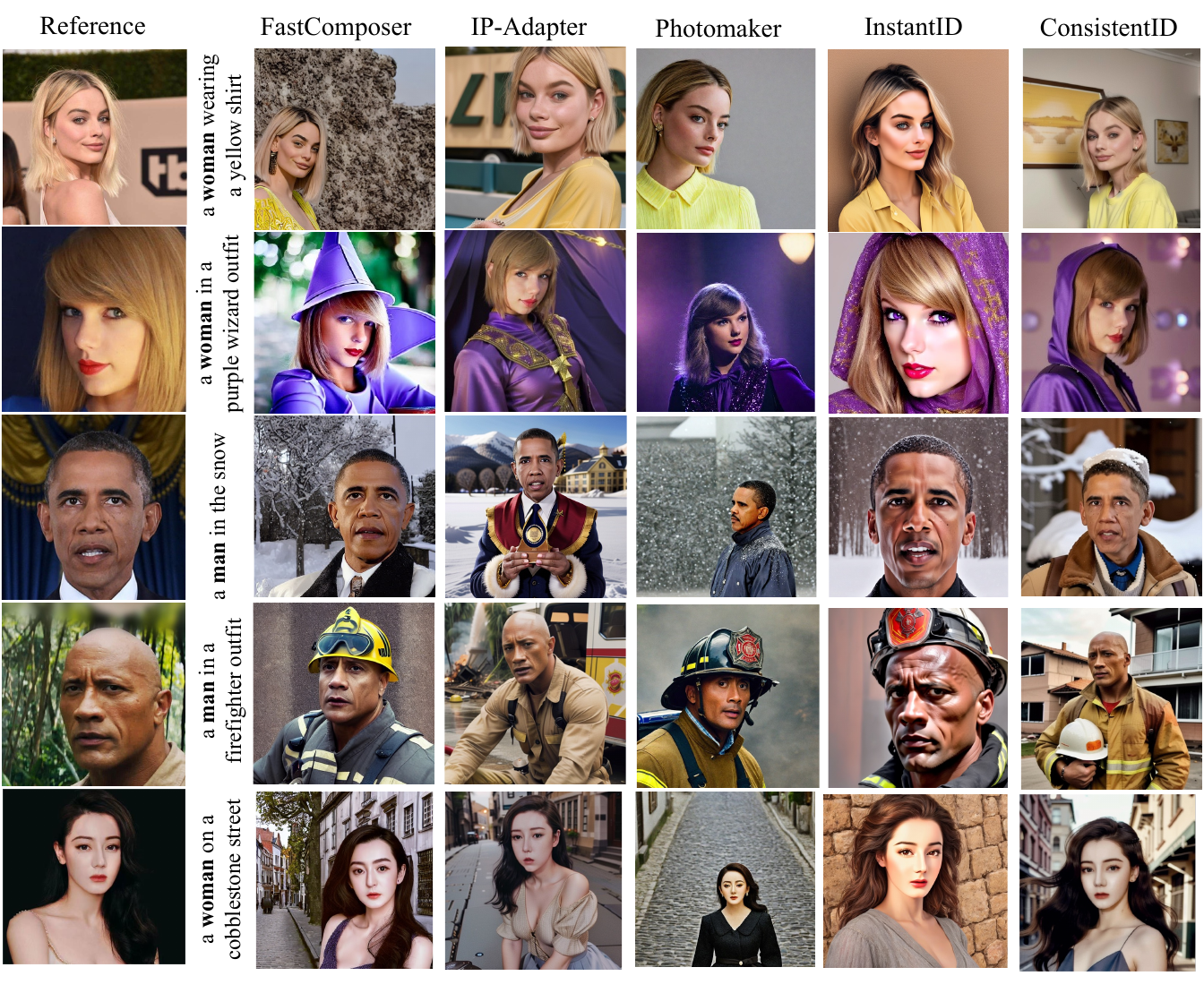}
  \caption{Qualitative comparison of universal recontextualization samples is conducted, comparing our approach with other methods using five distinct identities and their corresponding prompts.  
Our ConsistentID exhibits a more powerful capability in high-quality generation, flexible editability, and strong identity fidelity.} 
  \label{fig:result_compare}  
\end{figure*}

\noindent \textbf{Experimental Metrics.}  To evaluate the effectiveness and efficiency of ConsistentID, we employ six widely used metrics~\citep{ruiz2023dreambooth}: CLIP-I~\citep{gal2022image}, CLIP-T~\citep{radford2021learning}, DINO~\citep{cong2020dovenet}, FaceSim~\citep{schroff2015facenet}, FID~\citep{heusel2017gans}, and inference speed. 
CLIP-T measures the average cosine similarity between prompt and image CLIP embeddings, evaluating ID fidelity.  
CLIP-I calculates the average pairwise cosine similarity between CLIP embeddings of generated and real images, assessing prompt fidelity.  
DINO represents the average cosine similarity between ViT-S/16~\cite{dosovitskiy2020image} embeddings of generated and real images, indicating fine-grained image-level ID quality.  
FaceSim determines facial similarity between generated and real images using FaceNet~\citep{schroff2015facenet}.  
FID gauges the quality of the generated images~\citep{heusel2017gans}.  
Inference speed denotes the calculation time under the same running environment.  
To fully evaluate and quantify quality at the facial region level, we propose a novel metric, \textbf{FGIS} (Fine-Grained Identity Similarity). 
FGIS is computed as the average cosine similarity between DINO embeddings of the generated facial regions in reference and generated images.  
A higher FGIS value indicates greater ID fidelity in the generated facial regions.

 \begin{table*}[htp]
    \centering
    \scalebox{1}{
     \begin{tabular}{ccccccc|c}
\hline 
& CLIP-T \(\uparrow\) & CLIP-I \(\uparrow\) & DINO \(\uparrow\) & FaceSim \(\uparrow\) & FGIS \(\uparrow\) & FID \(\downarrow\) & Speed (s) \\
\hline 
Fastcomposer~\citep{xiao2023fastcomposer} & 27.8 & 67.0 & 68.4 & 75.2 & 77.7 & 372.8 & \textbf{10} \\
IP-Adapter~\citep{ye2023ip} & 27.6 & \underline{75.0} & 74.5 & 75.6 & 73.4 & 320.0 & 13 \\
Photomaker~\citep{li2023photomaker} & 30.7 & 71.7 & 72.6 & 69.3 & 73.2 & 336.5 & 17 \\
InstantID~\citep{wang2024instantid} & \underline{30.3} & 68.2 & \underline{77.6} & \underline{76.5} & \underline{78.3} & \textbf{271.9} & 19 \\
ConsistentID & \textbf{31.1} & \textbf{76.7} & \textbf{78.5} & \textbf{77.2} & \textbf{81.4} & \underline{312.4} & 16 \\
\hline 
\end{tabular}
}
    \caption{Quantitative comparison of the universal recontextualization setting on the MyStyle test dataset. The benchmark metrics assessed text consistency (CLIP-T), the preservation of coarse- and fine-grained ID information (CLIP-I, DINO, FaceSIM, and FGIS),  generation quality (FID), and inference efficiency (speed in seconds). }
    \label{tab:my_label}
\end{table*}

\subsection{Comparation Results}
\label{comparation_results}
To demonstrate the effectiveness of ConsistentID, we conduct a comparative analysis against state-of-the-art methods, including Fastcomposer~\cite{xiao2023fastcomposer}, IP-Adapter\cite{ye2023ip}, Photomaker~\cite{li2023photomaker}, and InstantID~\cite{wang2024instantid}.  
During testing, we utilized the officially provided models and generate portrait images using only a single reference image.
Consistent with the evaluation protocol of Photomaker, we utilize the Mystyle~\cite{nitzan2022mystyle} dataset for quantitative assessment and incorporate over ten identity datasets for qualitative visualization. 

\begin{figure}[tb]
  \centering  
  \includegraphics[width=0.5\textwidth]{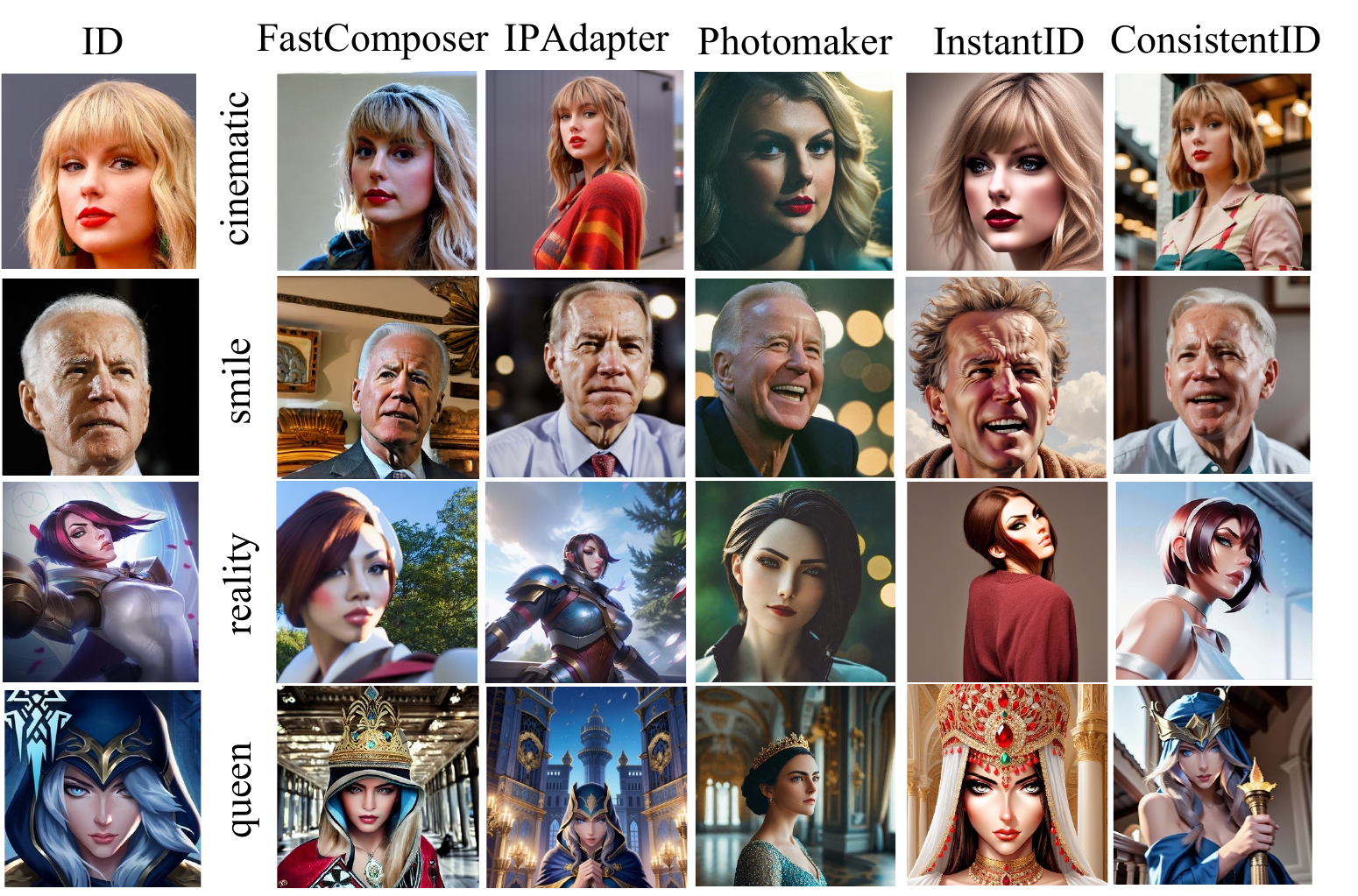}
  \caption{Qualitative comparison of our model with other models on the task of decoupling ID and text instructions.
  Notably, our method can control both the expression and style of the ID, and also demonstrates excellent results in maintaining the identity of anime characters.
 }
  \label{fig:facial_compare_1}  
\end{figure}

\begin{figure*}[tb]
  \centering  
  \includegraphics[width=1.0\textwidth]{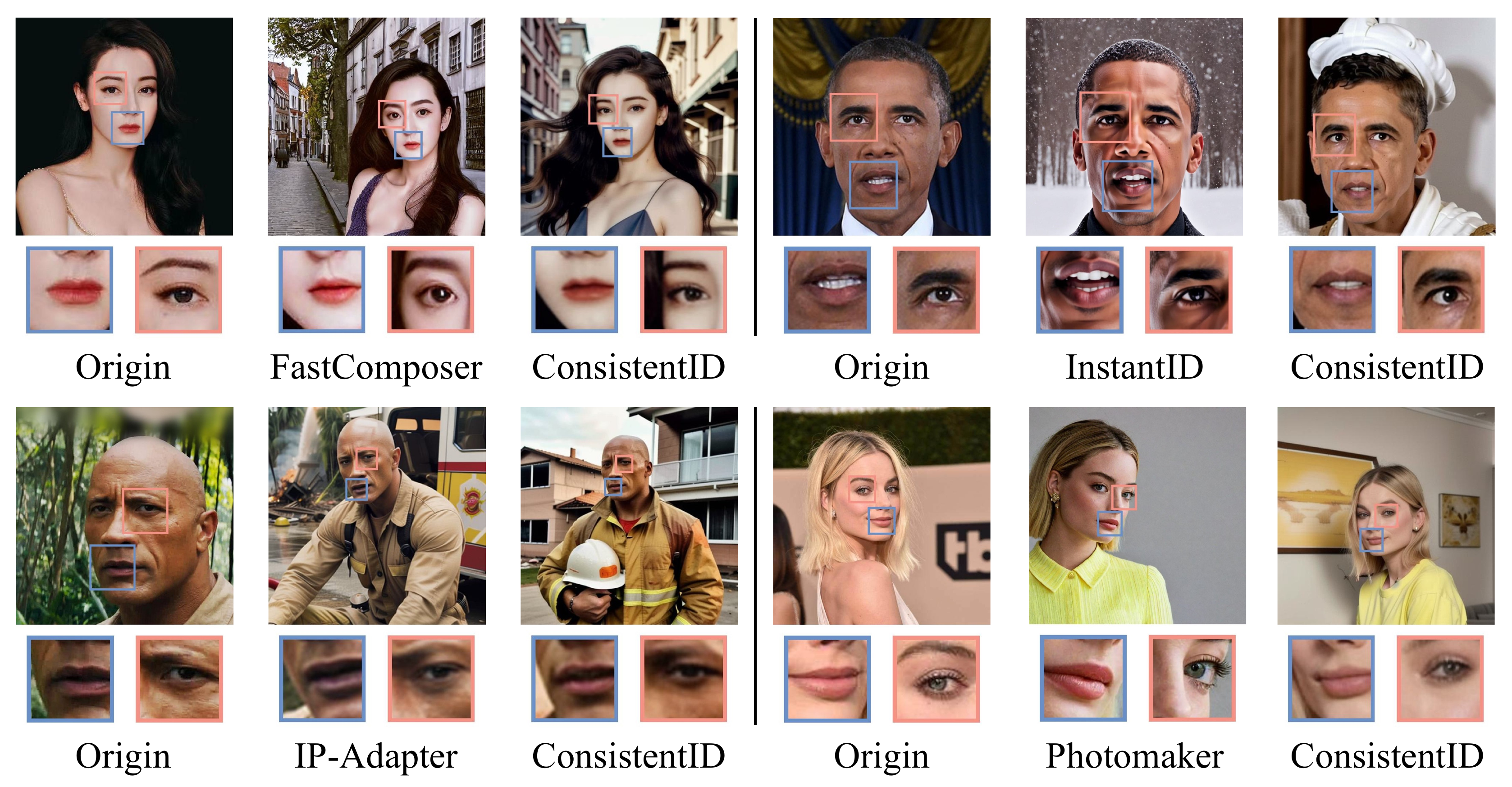}
  \caption{Comparison of facial feature details between our method and existing approaches. Notably, the characters generated by our method exhibit superior ID consistency in facial features such as eyes, nose, and mouth.}
  \label{fig:facial_compare}  
\end{figure*}

\noindent \textbf{Quantitative results:} 
Following Photomaker~\cite{li2023photomaker}, we used the test dataset from Mystyle~\cite{nitzan2022mystyle}. 
Notably, during the inference process, we did not use LLaVA to generate the descriptions of each facial region but obtained facial feature information using a fixed phrase: `This person has one nose, two eyes, two ears, and a mouth.' as the text prompt. 
The quantitative comparison is conducted under the universal recontextualization setting, utilizing a set of metrics to benchmark various aspects.  

The results are showcased in Table~\ref{tab:my_label}. 
A thorough analysis of the table demonstrates that ConsistentID consistently outperforms other methods across most evaluated metrics, and surpasses other IP-Adapter-based methods in terms of generation efficiency.  
This is attributed to ConsistentID's fine-grained ID preservation capability and the efficiency of the lightweight multimodal facial prompt generator. 
Regarding the FID metric, the relatively lower performance may be influenced by the inherent generative limitations of the base model SD1.5. 

\noindent \textbf{Visualized comparisons under different scenarios:} To demonstrate the advantages of ConsistentID visually, we present the text-edited generation results of all methods, using reference images from five distinct identities, in Figure~\ref{fig:result_compare}. 
This visualization highlights ConsistentID's capability to produce vibrant and realistic images, with a particular emphasis on facial features.  
To further elucidate this observation, we selectively magnify and compare specific facial details across all methods in four identities, as depicted in Figure~\ref{fig:facial_compare}.   
Our model showcases exceptional ID preservation capabilities in facial details, especially in the eyes and nose, attributed to fine-grained multimodal prompts and facial regions' ID information. 

\begin{figure*}[tb]
    \centering
    \includegraphics[width=1.0\linewidth]{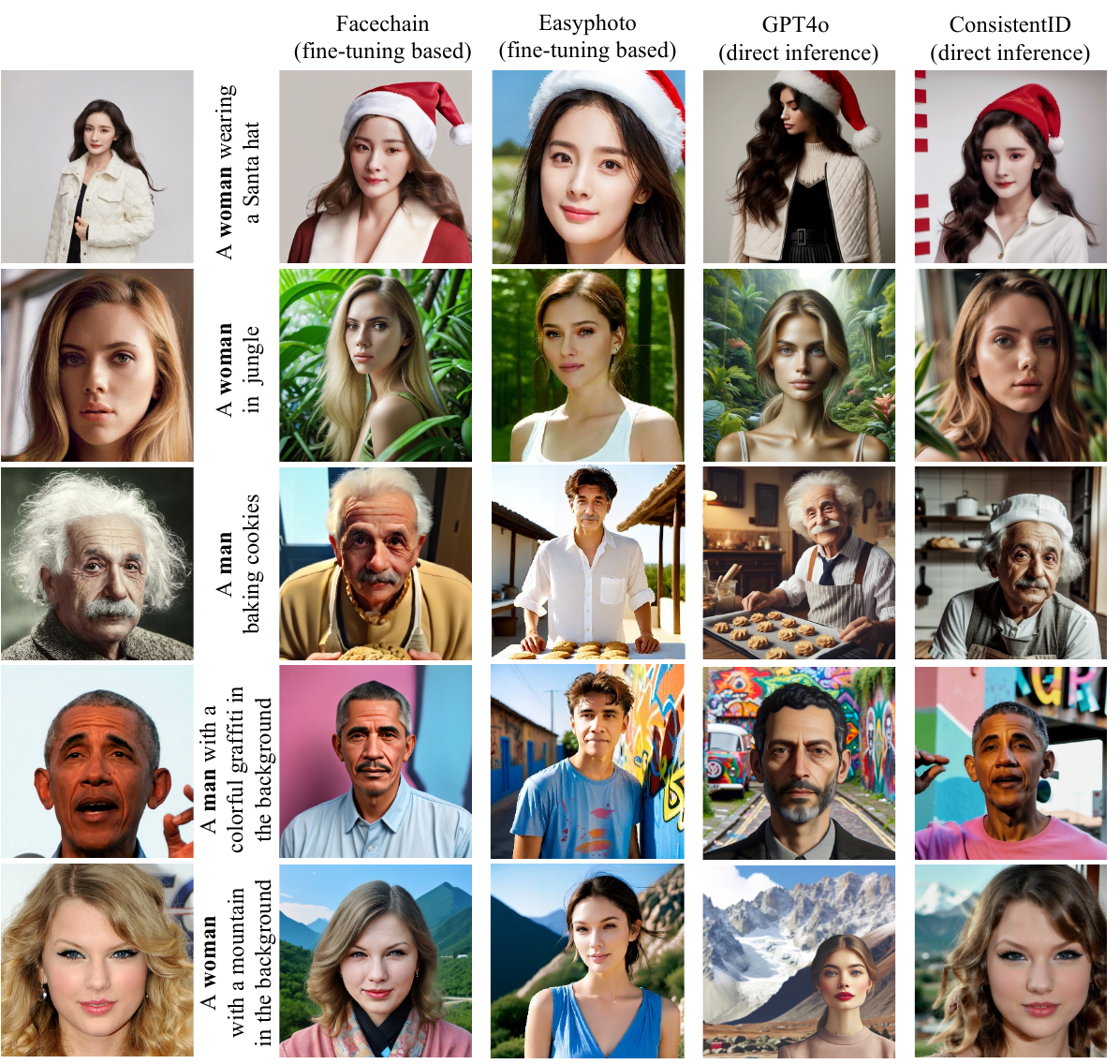}
    \caption{The comparisons with more fine-tuning-based and direct inference models. For direct inference method GPT4o, we use stars' names in the text prompt.}
    \label{fig:more_compare}
\end{figure*}

To validate the capability of accurate text understanding, we additionally present style-based and action-based text-edited results in Figure~\ref{fig:facial_compare_1}.  
It can be observed that the images generated by InstantID exhibit limited flexibility in facial poses. 
This limitation is likely attributed to Controlnet-based prompt insertion methods, which may easily overlook textual prompts. 
Simultaneously, we notice that, while Photomaker can accurately comprehend textual and visual prompts, it lacks the ID consistency of facial regions. 
In contrast, our ConsistentID achieves optimal generation results due to its precise understanding of textual and visual prompts.
This further emphasizes the importances of multimodal fine-grained ID information. 
To fully show the advantages of our ConsistentID, more visualized comparisons are provided in following section, including comparative experiments with fine-tuning-based models (Facechain~\citep{liu2023facechain}, and Easyphoto~\citep{wu2023easyphoto}), and the MLLM model GPT4o. 

\begin{figure*}[tb]
  \centering  
  \includegraphics[width=1.0\textwidth]{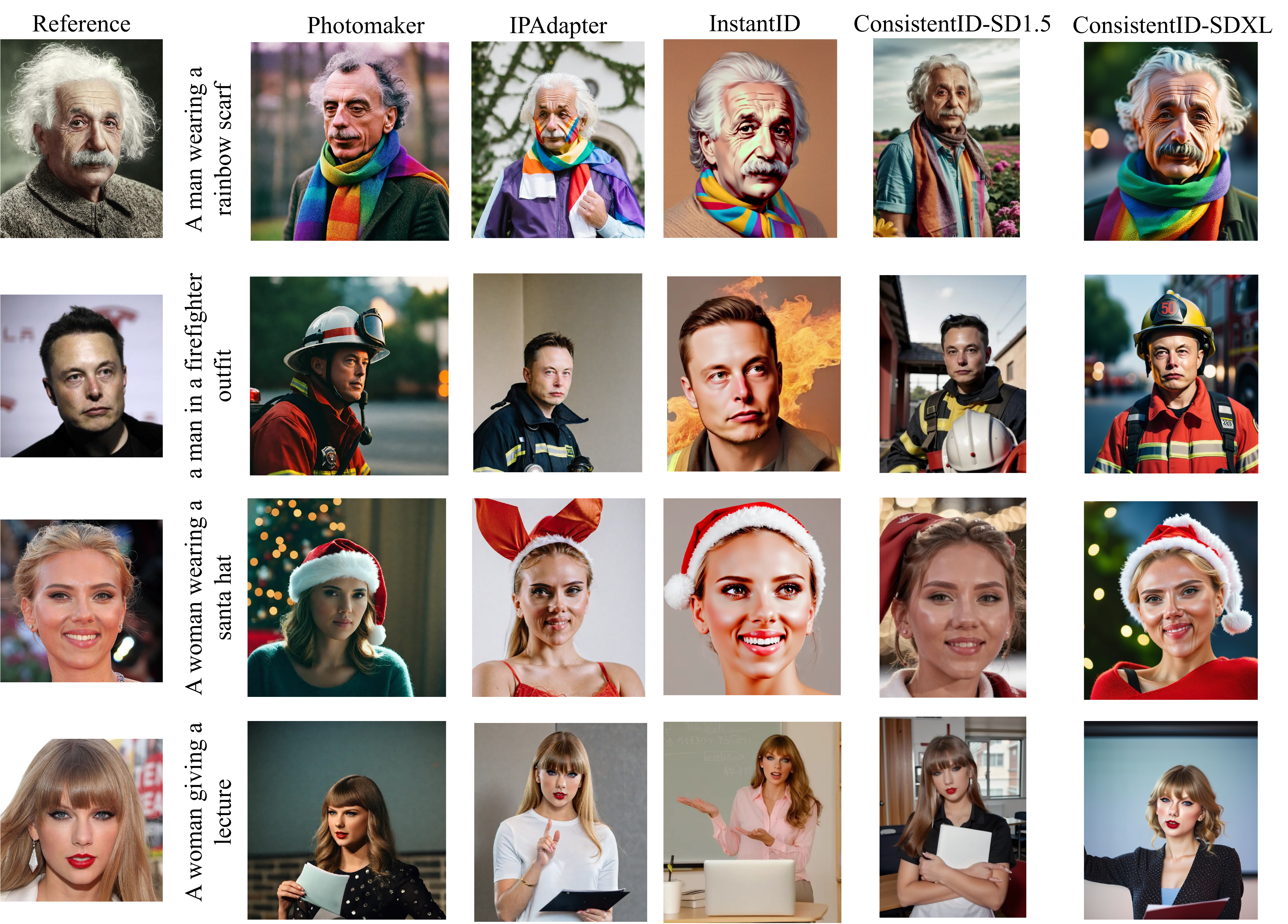}
  \caption{Results of our proposed ConsistentID model extended on the SDXL base model, compared with other SDXL-based models.} 
  \label{fig:SDXL}  
\end{figure*}

\subsection{Compare with more models}
\label{sec:fine-tuning}
To further illustrate the advantages of our model, we compare ConsistentID with two LoRA-based fine-tuning methods, Facechain~\citep{liu2023facechain} and Easyphoto~\citep{wu2023easyphoto}, as well as the closed-source model GPT4o~\footnote{https://chatgpt.com/}. 
To ensure a fair comparison, all models are provided with a single reference image as input.   
For the LoRA-based methods, the input comprises \textit{image} + \textit{prompt}, where \textit{image} represents the reference image and \textit{prompt} specifies the editing instruction for generating personalized outputs.  
For the GPT4o model, the template used is: `Edit this image using the following instruction: {person\_name} + \textit{prompt}', where {person\_name} specifies the individual's name. 
In Figure~\ref{fig:more_compare}, we show the visualized results of the personalization comparison using a single image, demonstrating the superior capability of our ConsistentID in keeping ID consistency. 

\begin{figure}[tb]
  \centering  
  \includegraphics[width=0.5\textwidth]{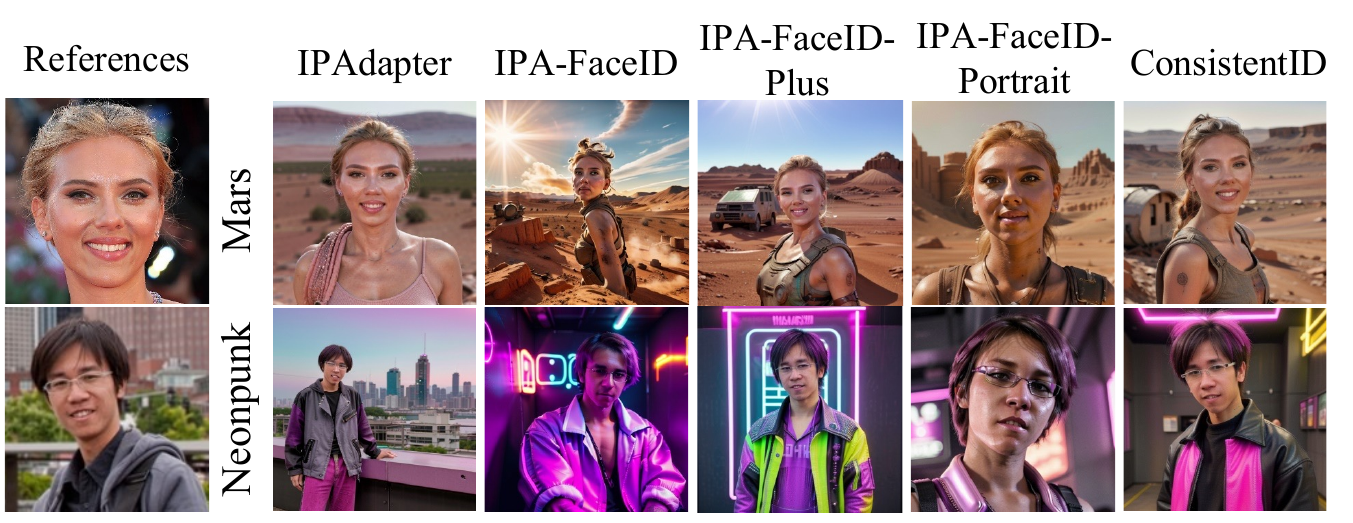}
  \caption{Comparison of ConsistentID with IP-Adapter and its face version variants conditioned on different styles.} 
  \label{fig:ipa_compare} 
\end{figure}

In addition, we compare ConsistentID with models specifically designed using IP-Adapter as the base model, as shown in Figure~\ref{fig:ipa_compare}. 
The figure clearly demonstrates that these models struggle to achieve detailed ID preservation in facial regions, primarily due to the lack of fine-grained textual and visual prompts. 
In contrast, ConsistentID exhibits a robust capability to preserve the integrity of facial ID and seamlessly blend it into various styles, utilizing multimodal fine-grained prompts. 
This highlights ConsistentID's advantage in retaining identity while providing enhanced flexibility and control over style customization.

\subsection{Human Study}
\label{human_study}

To accurately capture user preferences, we conducted surveys to evaluate perceptions of image fidelity, fine-grained ID fidelity, and overall ID fidelity.  
Figure~\ref{fig:user_pref} visualizes the proportion of total votes received by each method.  
Across all three metric dimensions, ConsistentID achieves the highest user preference share.  
This result highlights the effectiveness of ConsistentID in meeting user expectations for both identity preservation and image quality.

\subsection{Ablation Study}
\noindent \textbf{Facial ID Types:} 
We conducted an ablation study on facial IDs, considering three variations: using only the holistic facial ID, using our designed fine-grained IDs, and using both the holistic facial ID and our designed fine-grained IDs. 
The results in Table~\ref{tab:aba_id} indicate that using fine-grained ID features alone effectively maintains ID consistency in the facial region, particularly evidenced by an increase in DINO values. 
In contrast, using only overall facial features (FaceID) results in decreased values across all metrics, due to the coarse-grained characteristic of the overall facial feature.  
Finally, the combination of holistic facial ID and fine-grained ID features achieves the best overall ID fidelity, demonstrating the complementary strengths of both approaches and their contribution to enhanced model performance. 

\begin{table*}[htb]
    \centering
    \begin{minipage}{1.0\linewidth}
    \centering
    \scalebox{1}{ 
\begin{tabular}{cccc|c|ccc|c} 
\hline 
\makebox[0.1\textwidth][c]{\text{\(\mathcal{L}_{noise}\)}} & \makebox[0.1\textwidth][c]{\text{\(\mathcal{L}_{facial}\)}} & \text{CLIP-I} \(\uparrow\) & \text{DINO} \(\uparrow\) & \text{FGIS} \(\uparrow\) & LLaVA1.5 & \text{CLIP-I} \(\uparrow\) & \text{DINO} \(\uparrow\) & \text{FGIS} \(\uparrow\) \\
\hline
\checkmark & $-$ & 66.4 & 77.8 & 82.9 & $-$ & 67.4 & 84.3 & 84.9 \\
\checkmark & \checkmark & \textbf{75.5} & \textbf{86.1} & \textbf{85.6} & 
\checkmark & \textbf{75.3} & \textbf{85.4} & \textbf{85.8} \\
\hline
\hline 
\multicolumn{2}{c}{Facial Feature} & CLIP$-$I \(\uparrow\) & DINO \(\uparrow\) & FGIS \(\uparrow\) & ImageProjection & \text{CLIP$-$I} \(\uparrow\) & \text{DINO} \(\uparrow\) & \text{FGIS} \(\uparrow\) \\
\hline 
\multicolumn{2}{c}{Overall Facial Feature} & 72.9 & 80.7 & 84.2 & $-$ & 61.0 & 75.6 & 82.9 \\
\multicolumn{2}{c}{Fine-grained Feature} & 75.2 & \textbf{86.6} & 85.4 & \checkmark & \textbf{75.5} & \textbf{86.1} & \textbf{85.6} \\
\multicolumn{2}{c}{Overall Facial \& Fine-grained Feature} & \textbf{75.5} & 86.1 & \textbf{85.6} & & & & \\
\hline
    \end{tabular}
}
    \end{minipage}
    \caption{Ablation study on ID features, loss functions, ImageProjection module, and the usage of LLaVA1.5 in training. } 
    \label{tab:aba_id}
\end{table*}

\begin{figure}[tb]
  \centering  
  \includegraphics[width=0.5\textwidth]{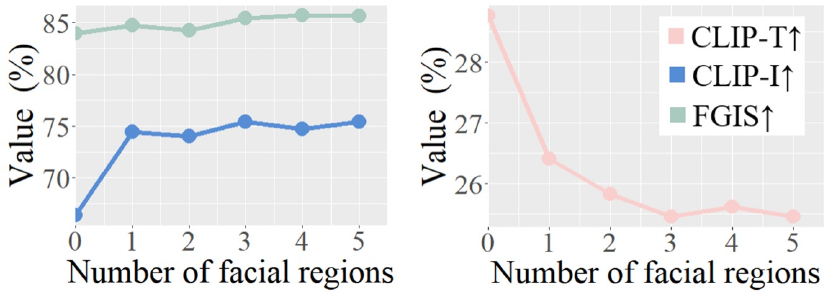}
  \caption{Ablation study of the number of facial regions.}
  \label{fig:aba_region}  
\end{figure}

\begin{figure}[tb]
  \centering  
  \includegraphics[width=0.5\textwidth]{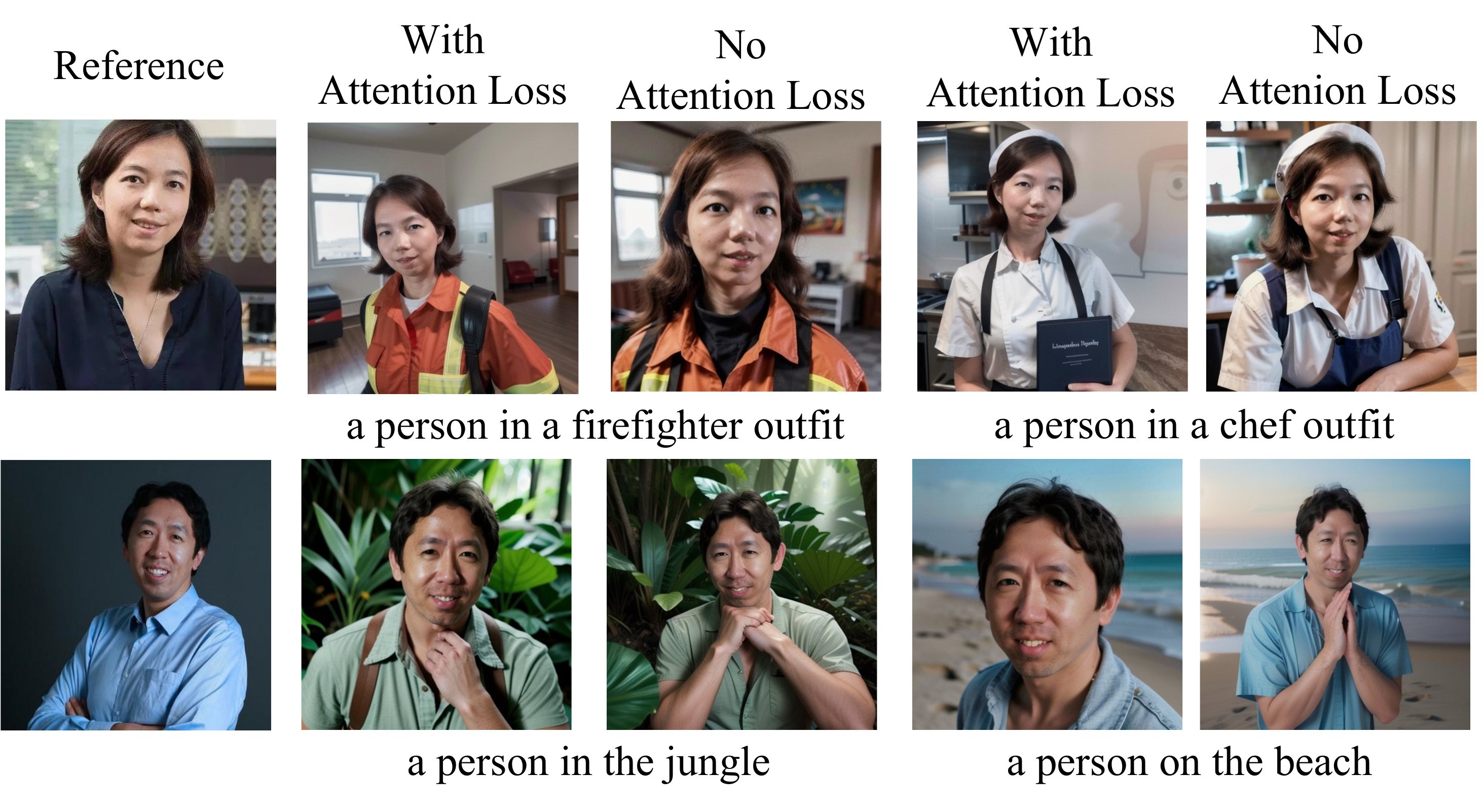}
  \caption{Visualized results with or without using attention loss.}
  \label{fig:loss_attenion}  
\end{figure}

\begin{figure}[tb]
  \centering  
  \includegraphics[width=0.5\textwidth]{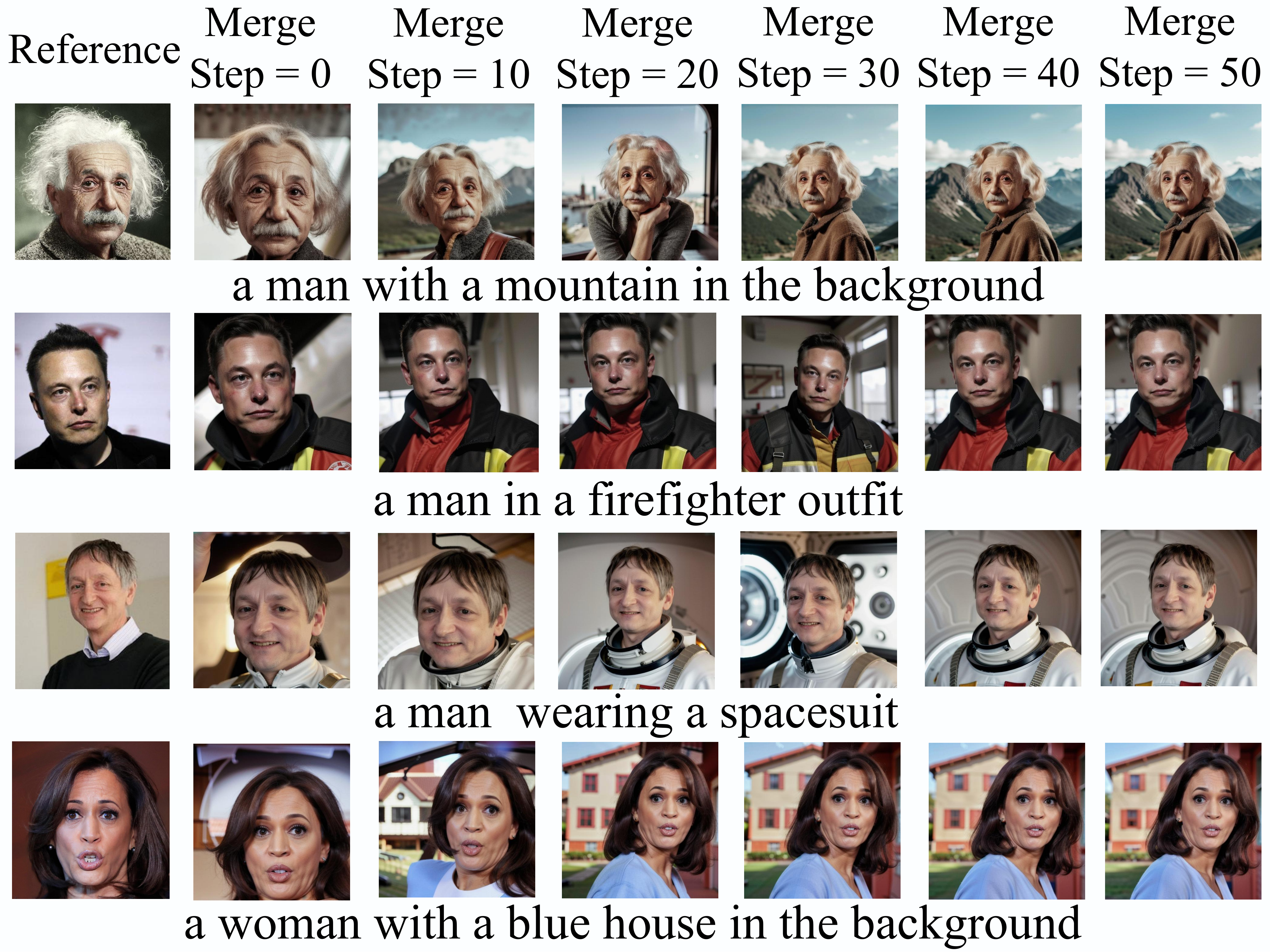}
  \caption{Visualized results under different `merge steps'. `Merge Step' indicates when to start adding facial image features to the text prompt.}
  \label{fig:vis_delay_control}  
\end{figure}

\begin{figure}[H]
  \centering  
  \includegraphics[width=0.5\textwidth]{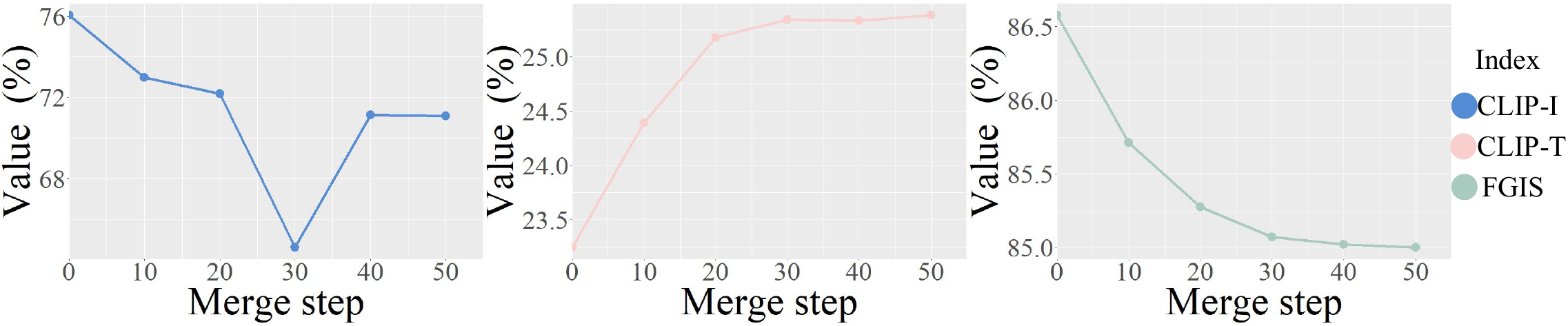}
  \caption{Performance variations of CLIP-I, CLIP-T, and FGIS metrics with increasing `merge step'.}
  \label{fig:delay_control}  
\end{figure}

\noindent \textbf{Facial attention localization strategy:} We investigated the effectiveness of facial attention localization strategies during training.  
The first strategy involves $\mathcal{L}_{noise}$, while the second strategy adds attention loss $\mathcal{L}_{facial}$.   
From Table~\ref{tab:aba_id}, we observe that ConsistentID experiences a clear improvement in metrics related to facial feature consistency and fine-grained ID preservation when $\mathcal{L}_{facial}$ is considered.  
This confirms the effectiveness of maintaining ID consistency between facial regions and the entire face during the training process. 

\noindent \textbf{Image projection module:} Additionally, we compared two training strategies.  
The first involves frozen weights of the image projection model and only training our designed FacialEncoder.  
The second strategy involves training both simultaneously. The results from Table~\ref{tab:aba_id} indicate that concurrently training ImageProjection brings the maximum benefits to the model. 
This is attributed to our model being an ID preservation method of cooperative training with multimodal text and image information. 

\noindent \textbf{The LLaVA1.5 usage:} 
In Table~\ref{tab:aba_id} (bottom), we analyze the impact of using LLaVA 1.5 to generate textual descriptions during training.  
Our observations indicate that incorporating LLaVA 1.5 leads to noticeable improvements across various metrics, with particularly significant gains in the CLIP-I metric.  
Notably, during inference, our model achieves satisfactory performance without relying on LLaVA 1.5, using only the simplified facial description, `This person has one nose, two eyes, two ears, and a mouth,' instead of generating detailed facial region descriptions. 
This demonstrates that our model effectively learns fine-grained features during the training phase with LLaVA 1.5, enabling it to maintain strong image understanding and identity preservation during inference.

\noindent \textbf{Different Facial Areas' Number:} To explore the influence of different numbers of facial areas, we adhere to the sequence `face, nose, eyes, ears, and mouth' and incrementally introduce the selected facial areas, as depicted in Figure~\ref{fig:aba_region}.   
 We note a progressive enhancement in image quality as the number of facial regions increases, attributed to the richer multi-modal prompts.  
 However, with regard to the CLIP-T metric, detailed textual descriptions encompass a greater variety of objects, potentially leading to oversight by the CLIP model.

\noindent \textbf{Attention loss $\mathcal{L}_{facial}$:} 
To further validate the effectiveness of our $\mathcal{L}_{facial}$, we present several comparisons using two different identities in Figure~\ref{fig:loss_attenion}. 
From the figure, we draw the following conclusions:
1) The details of key facial areas, such as Taylor's eye shape, are well preserved. 
2) The appearances of facial regions, such as eyes, ears, and mouths, effectively reflect identity consistency with the reference images.

\begin{figure}[tb]
  \centering  
  \includegraphics[width=0.5\textwidth]{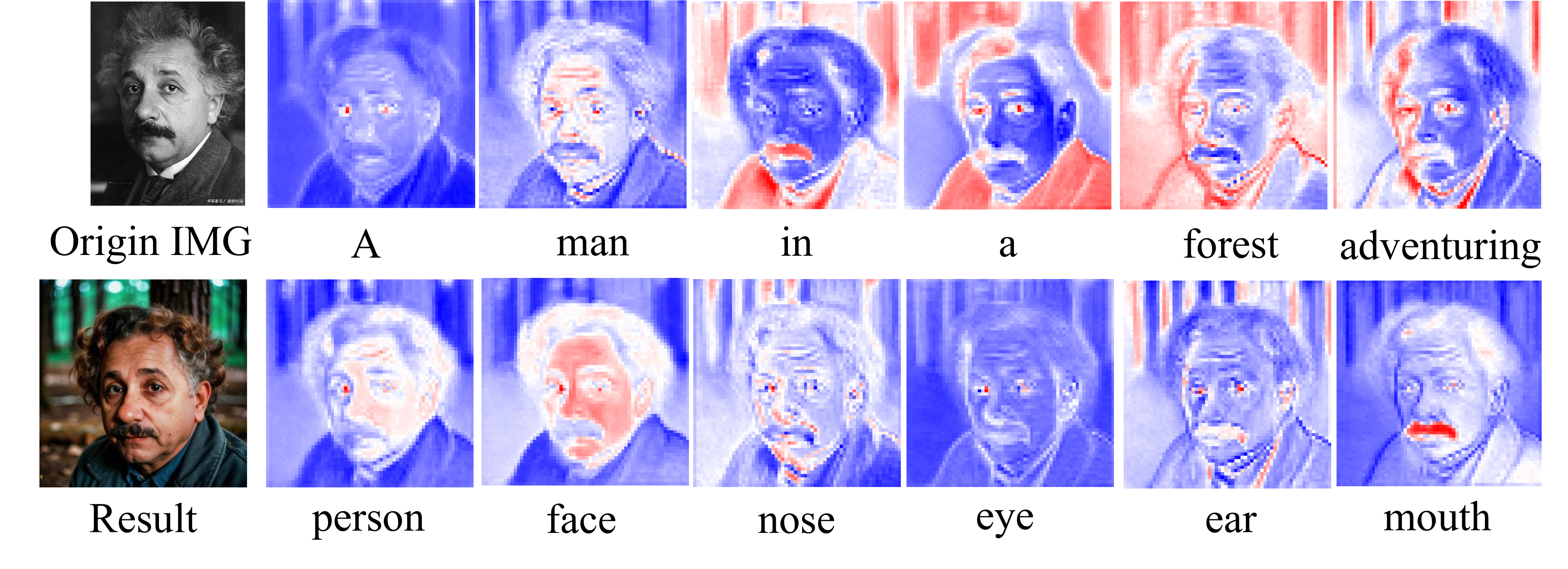}
  \caption{Experiments to visualize facial changes in attention map during training results.}
  \label{fig:aba_attention}  
\end{figure}

Additionally, as shown in Figure~\ref{fig:aba_attention}, the attention scores in the model's facial feature regions gradually increase as training progresses. 
This indicates improved semantic alignment between the facial regions and their corresponding textual descriptions, further supporting the efficacy of $\mathcal{L}_{facial}$. 

\noindent \textbf{Delay control:} 
In Figure~\ref{fig:vis_delay_control}, we provide visualized results to evaluate the impact of delay control during inference.  
The term `merge step' denotes the first time step in which we incorporate fine-grained facial image features. It helps to control the balance between text prompts and face images. In general, as the `merge step' increases, the influence of fine-grained facial image features gradually diminishes. 

For example, if the `merge step' is set to 0, it indicates that fine-grained facial image features are dominant in the generation process and might result in semantic inconsistencies. On the contrary, setting the `merge step' to 0 will maximize the guidance of text prompts, yet might harm the identity consistency.

To visually illustrate the impact of the `merge step', we display the variation curves for the CLIP-I, CLIP-T, and FGIS metrics as the `merge step' increases in Figure~\ref{fig:delay_control}. 
From the figure, we observe a consistent trend where the textual control gradually strengthens with each increment of the `merge step'. 

\noindent \textbf{Inference time for each module:} In Table~\ref{tab:time_consumption}, we show the processing time for each module involved in handling a single image during inference, with a total duration of 16 seconds per image.   
This quantitative analysis highlights the competitive performance of the system in terms of inference efficiency \footnote{Notably, during the inference phase, the usage of the LLaVa or ChatGPT-4v module is not mandatory. Satisfactory results can be obtained by relying exclusively on predefined facial features prompts, such as `face, nose, eyes, ears, and mouth.'}.  

\noindent \textbf{Extensibility of methods for base models:} We further replace the base model in the ConsistentID framework with the SDXL model, as illustrated in Figure~\ref{fig:SDXL}. 
 From the figure, ConsistentID effectively enhances facial feature consistency and outperforms other models in terms of ID similarity.   
Furthermore, we conducted additional ablation experiments using the SDXL model as the base for the ConsistentID framework, as shown in Table~\ref{tab:aba_SDXL}.  
Compared to the SD1.5-based ConsistentID model, the SDXL-based version demonstrates notable improvements across key metrics, including facial similarity, CLIP-I, CLIP-T, and FID.  
These results further validate the robust generalization capabilities of the ConsistentID model and demonstrate its flexibility in utilizing more advanced base models for improved performance. 

\begin{table}[h]
    \centering
    \Huge
    \resizebox{0.5\textwidth}{!}{
    \begin{tabular}{lcccc|c}
\hline & \text { BiSeNet } & \text { InsightFace } & \text { FacialEncoder } & \text { UNet }& \text { Inference } \\
\hline \text { Time (s) } & 1 & 3 & 3 & 5 & 4\\
\hline
    \end{tabular}
    }
    \caption{Inference time of each module.} 
    \label{tab:time_consumption}
\end{table}

\begin{table}[h]
    \centering
    \resizebox{0.5\textwidth}{!}{
    \begin{tabular}{lccc|c}
\hline 
& \text{CLIP-T} \(\uparrow\) & \text{FaceSim} \(\uparrow\) & \text{FID} \(\downarrow\) & \text{Speed (s)} \(\downarrow\) \\
\hline
\text { ConsistentID-SD1.5 } & 19.8 & 62.9 & 307.9 & \textbf{16} \\
\hline \text { ConsistentID-SDXL } & \textbf{21.9} & \textbf{73.1} & \textbf{304.5} & 18 \\
\hline
    \end{tabular}
    }
    \caption{ConsistentID extension ablation experiment, evaluating the performance using different base models. The average performance is tested on 180 images under 45 prompt conditions, using the character ID shown in Figure~\ref{fig:SDXL}.} 
    \label{tab:aba_SDXL}
\end{table}

\section{Conclusion}
\label{conclus_limi}
In this work, we introduce ConsistentID, an innovative method designed to maintain identity consistency and capture diverse facial details.  
We have developed two novel modules: a multimodal facial prompt generator and an identity preservation network.  
The former is dedicated to generating multimodal facial prompts by incorporating both visual and textual descriptions at the facial region level.  
The latter aims to ensure ID consistency in each facial area through a facial attention localization strategy, preventing the blending of ID information from different facial regions.  
By leveraging multimodal fine-grained prompts, our approach achieves remarkable identity consistency and facial realism using only a single facial image.  
Additionally, we present the FGID dataset, a comprehensive dataset containing fine-grained identity information and detailed facial descriptions essential for training the ConsistentID model.   
Experimental results demonstrate outstanding accuracy and diversity in personalized facial generation, surpassing existing methods on the MyStyle dataset.


\bibliographystyle{IEEEtran}
\bibliography{References}

\end{document}